\documentclass[11pt]{article}

\usepackage[preprint]{acl}
\usepackage{times}
\usepackage{latexsym}
\usepackage[T1]{fontenc}
\usepackage[utf8]{inputenc}
\usepackage{microtype}
\usepackage{graphicx}
\graphicspath{{figs/}}
\usepackage{amsmath}
\usepackage{amssymb}
\usepackage{booktabs}
\usepackage{multirow}
\usepackage{enumitem}
\usepackage{float}
\usepackage{tcolorbox}
\usepackage{pifont}

\floatstyle{ruled}
\newfloat{algorithm}{tbp}{loa}
\floatname{algorithm}{Algorithm}

\newcounter{algline}
\newcommand{\algline}[2][0]{%
  \par\noindent
  \hangindent=\dimexpr 2.8em + #1em\relax
  \hangafter=1
  \makebox[2.4em][r]{\stepcounter{algline}\arabic{algline}:}%
  \hspace*{0.4em}\hspace*{#1em}#2\par
}

\title{Model What Matters: Decision-Guided Selective World Modeling for Language Agents}
\author{Maolin Ran$^{1}$, Minda Chen$^{2}$, Di Huang$^{2}$, Wei Xiong$^{2}$, Fei Huang$^{2,*}$, Weiwen Liu$^{1,*}$ \\
$^{1}$Shanghai Jiao Tong University \quad $^{2}$LongShine AI Research \\
\texttt{\{maolinr03, wwliu\}@sjtu.edu.cn, huangfei@longshine.com}
}

\begin{document}
\maketitle

\begin{abstract}
Reinforcement learning provides a direct mechanism for optimizing task objectives in large language model-driven agents, but task rewards are typically sparse trajectory-level signals that only evaluate task success, without directly supervising action-conditioned environment transitions, leaving the model with limited explicit understanding of action consequences and environment dynamics. Recent work introduces auxiliary supervision through policy and world modeling co-training, yet typically treats complete environmental observations uniformly as world modeling targets, overlooking the varying decision relevance of observation content. We propose \textbf{DGS-WM} (\textbf{D}ecision-\textbf{G}uided \textbf{S}elective \textbf{W}orld \textbf{M}odeling), a decision-guided policy-world modeling co-training approach for language agents. DGS-WM employs a Decision-Guided Observation Selection component that filters observation units with decision value as world modeling targets based on their support for subsequent policy responses, and a Confidence-Adaptive Weighting component that adaptively adjusts supervision strength according to the model's prediction confidence on selected targets. Together, these components form a decision-aware auxiliary world modeling loss, jointly optimized with the RL loss on the same policy model. Across multiple benchmarks covering single-hop/multi-hop question answering and interactive web tasks, DGS-WM achieves consistent improvements under different model scales and RL algorithms, demonstrating more stable and effective performance compared to full-observation world modeling methods.
\end{abstract}

\section{Introduction}

Large language model (LLM)-driven agents can complete tasks such as retrieval, question answering, and web operations through multi-turn interactions with environments. Reinforcement learning (RL) provides a direct training mechanism for optimizing task objectives for such agents, enabling models to progressively improve action selection and task execution capabilities based on environmental observations and task rewards. In recent years, PPO, GRPO, and related methods have been widely applied to LLM reasoning optimization and agent training, achieving significant progress \citep{ppo,rlhf,deepseek-r1}.

Multi-turn tasks such as retrieval, question answering, and web operations typically exhibit long-horizon dependencies: current actions change environment states, and state changes determine subsequent available environmental observations and feasible actions \citep{gigpo,search-r1}. To make effective decisions continuously in such tasks, policies need to form usable representations of action consequences and environment dynamics. RL primarily optimizes policy behavior through task rewards; trajectory-level or outcome-level rewards typically only evaluate task success and do not directly supervise action-conditioned environment transitions \citep{search-r1,gigpo}. Therefore, models can learn which behaviors are more likely to obtain high rewards, but may not explicitly learn what environmental observations actions will produce and how these environmental observations affect subsequent interactions; the action-environmental observation relationships already recorded in rollouts are also not fully converted into independent training signals.

\begin{figure}[t]
\centering
\includegraphics[width=\columnwidth]{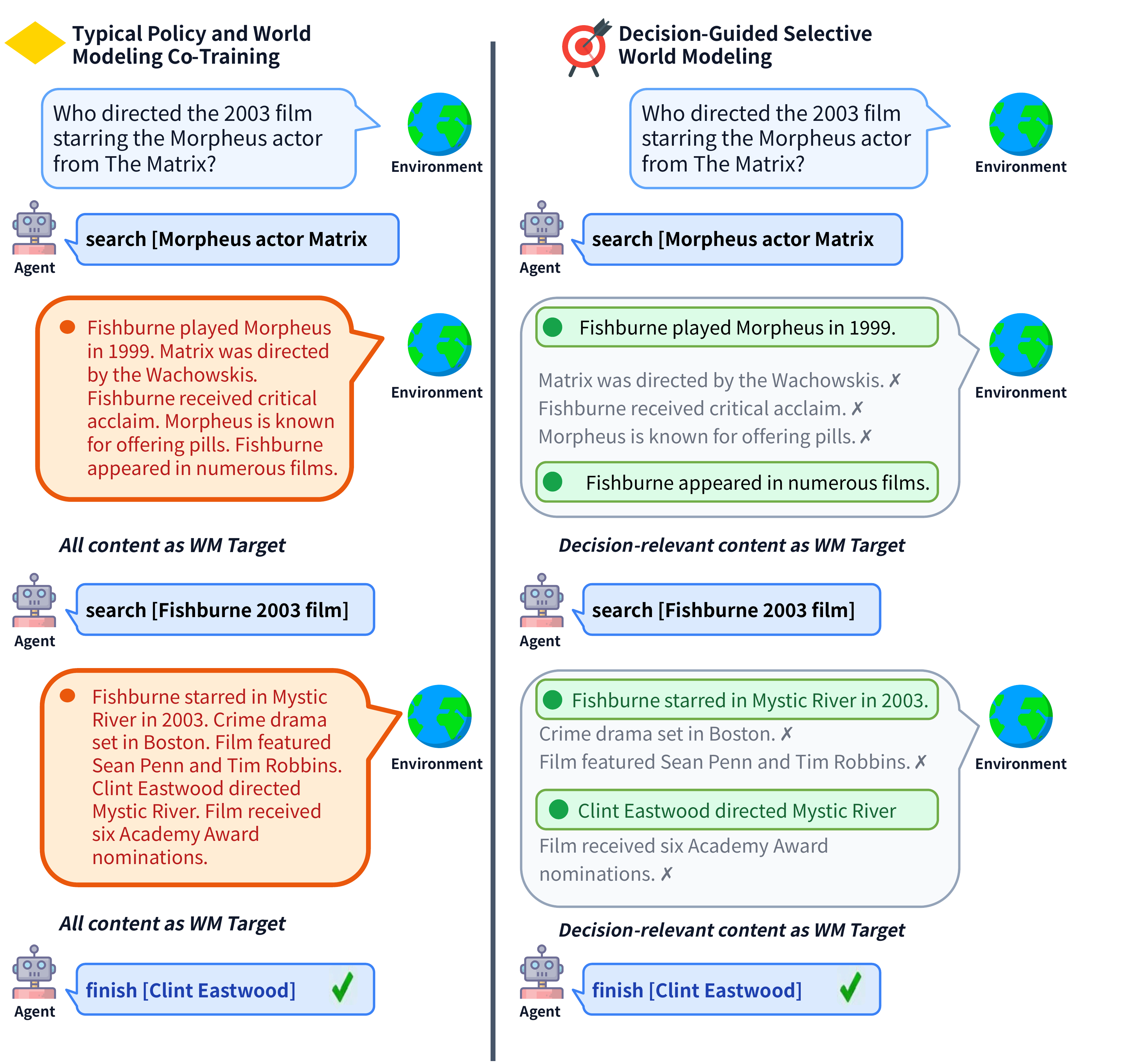}
\caption{Comparison of typical policy-world modeling co-training versus DGS-WM on a two-hop question answering task. Given the same environment observations, typical co-training methods supervise all observation content equally as WM targets (left), while DGS-WM selectively supervises only decision-relevant content based on their support for future policy responses (right). Green highlights indicate selected content; gray text indicates filtered noise.}
\label{fig:intro}
\end{figure}

World modeling methods provide explicit environment dynamics supervision for policies by learning subsequent states or textualized environmental observations conditioned on actions, thereby supplementing training signals that rely solely on task rewards. Recent research further combines this learning objective with on-policy RL: ECHO optimizes the policy objective while using environmental observations produced by actions in rollouts for auxiliary prediction; PaW further filters transitions to be included in world modeling training on this basis. For transitions included in joint training, both typically use their complete next observation as the world modeling target (Figure~\ref{fig:intro} left), enabling the policy model to learn environment dynamics during interactions beyond task rewards \citep{echo,paw}. This raises the question we focus on in this paper: within a single environmental observation, should the complete observation be treated by default as a unified world modeling target?

Our core insight is that environmental observations do not inherently possess equal supervision value. An environmental observation may simultaneously contain key information that can influence future decisions or content unrelated to the current task; if the complete environmental observation is uniformly used as a supervision target, the model must bear the same auxiliary prediction pressure for this content. Therefore, we argue that only observation content valuable for future decisions should be retained as world modeling supervision targets.

Based on this insight, we propose DGS-WM (Decision-Guided Selective World Modeling), a decision-guided policy-world modeling co-training method for text agents (Figure~\ref{fig:intro} right). DGS-WM first filters content with decision value from environmental observations through Decision-Guided Observation Selection and constructs them as world modeling targets; subsequently through Confidence-Adaptive Weighting, it adjusts auxiliary supervision strength based on the model's current prediction confidence for retained targets. Based on selected prediction targets, DGS-WM constructs world modeling objectives to jointly train the policy model together with RL objectives.

 DGS-WM first filters content with decision value from environmental observations through Decision-Guided Observation Selection and constructs them as world modeling targets; subsequently through Confidence-Adaptive Weighting, it adjusts auxiliary supervision strength based on the model's current prediction confidence for retained targets. Based on selected prediction targets, DGS-WM constructs world modeling objectives to jointly train the policy model together with RL objectives.

The main contributions of this paper are as follows:

\begin{enumerate}
    \item We propose DGS-WM, an RL and world modeling co-training algorithm for text agents that implements decision-oriented observation selection and adaptive supervision strength control through Decision-Guided Observation Selection and Confidence-Adaptive Weighting.
    \item We propose a core insight: the world modeling value of environmental observations is not uniform, and complete observations should not be treated by default as unified supervision targets. Based on this, we refine supervision granularity to within observations, retaining only content valuable for future decisions as world modeling targets, and adjusting auxiliary supervision strength through confidence-adaptive weighting.
    \item We conduct systematic evaluation on multiple public benchmarks covering single-hop/multi-hop question answering and web interaction. Results show that DGS-WM achieves consistent improvements across different tasks, model scales, and RL algorithms; in contrast, full observation world modeling methods (ECHO) show unstable performance in some settings, sometimes underperforming pure RL baselines.
\end{enumerate}

\section{Related Work}

\subsection{Reinforcement Learning for LLM Agents}

Reinforcement learning has become an important paradigm for optimizing LLM policies. Existing methods first revolve around policy updates and advantage estimation: PPO \citep{ppo} constrains on-policy updates with clipped surrogate objectives, REINFORCE-style methods \citep{reinforce} and RLOO \citep{rloo} reduce reliance on explicit value models through sequence-level returns and leave-one-out baselines, while GRPO \citep{grpo} performs group-based optimization using relative returns from multiple samples under the same input. Subsequently, extensive work has improved the training efficiency and stability of LLM RL from aspects such as clipped objectives, sampling strategies, loss aggregation, and importance weights, such as DAPO \citep{dapo} and GMPO \citep{gmpo}. For long-sequence and multi-turn agents, research further focuses on how to perform finer-grained credit assignment using sparse outcome rewards, including GiGPO \citep{gigpo} based on episode/state grouping, SPO \citep{spo} based on segment-level advantage, and Tree-GRPO \citep{tree-grpo} that constructs intermediate process signals using structured rollouts. Compared to outcome-level rewards, process supervision provides finer-grained training signals for intermediate steps, but typically involves additional annotation, estimation, or data construction costs \citep{process-supervision}. The method proposed in this paper provides auxiliary supervision for RL training through world modeling, requiring no additional annotations and can be combined with different on-policy RL algorithms (such as GRPO).

\subsection{World Modeling for LLM Agents}

World modeling methods for language agents mainly revolve around learning action-conditioned environment transitions, i.e., predicting subsequent states or textualized environmental observations based on interaction history and actions. Early work typically places the learned world model outside the policy optimization process, serving as an environment simulator or planning module for generating imagined rollouts, performing search, or conducting lookahead planning \citep{world-models,reasoning-via-planning,imagine-then-plan}. Building on this, the WM-to-RL staged paradigm first uses interaction trajectories to learn environment dynamics, future state prediction, or other environment knowledge, then uses the obtained world model checkpoint or representations for subsequent policy optimization; Early Experience \citep{early-experience} and RWML \citep{rwml} represent the reward-free world model warm-up and the approach of first performing world model learning then task-success RL, respectively. More recent work places world model learning and on-policy policy optimization in the same training process, forming a policy-world modeling co-training paradigm \citep{echo,paw}. Under this paradigm, environmental feedback included in WM training typically directly uses the complete next observation as the world modeling target, without further distinguishing content within the feedback, making all content jointly bear auxiliary supervision. DGS-WM continues this co-training paradigm but consciously selects world modeling targets, retaining only feedback content valuable for future decisions for world modeling supervision.

\section{Methodology}

\begin{figure*}[t]
\centering
\includegraphics[width=\textwidth]{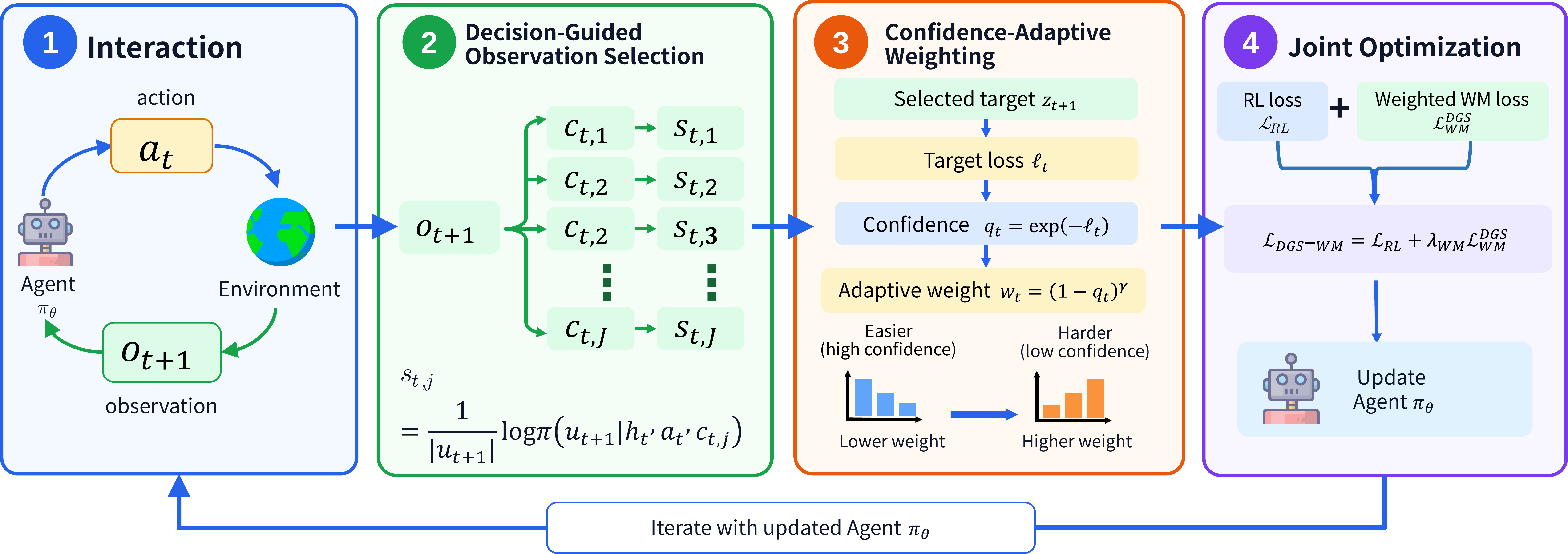}
\caption{Overview of DGS-WM. Complete trajectories provide RL signal; observations and subsequent responses are used for target selection; history, action, and selected targets are used for world modeling prediction, producing target-level confidence and weights; weighted world modeling loss jointly updates policy with RL loss.}
\label{fig:method}
\end{figure*}

We propose \textbf{DGS-WM} (\textbf{D}ecision-\textbf{G}uided \textbf{S}elective \textbf{W}orld \textbf{M}odeling), a policy-world modeling co-training method for language agents that introduces selective environmental observation prediction supervision during RL training. As illustrated in Figure~\ref{fig:method}, \textbf{Decision-Guided Observation Selection} uses subsequent policy responses to select world modeling targets from observations, and \textbf{Confidence-Adaptive Weighting} adjusts supervision strength based on the model's prediction confidence for selected targets. The resulting world modeling loss is jointly optimized with the RL loss to update the same policy model.

\subsection{Problem Formulation and Auxiliary World Modeling}

In text interaction tasks, a language model policy $\pi_\theta$ generates responses and executes actions $a_t$ based on interaction history $h_t$ (containing task instructions and prior interactions). After receiving environmental observation $o_{t+1}$, the policy generates the next complete response:

\begin{equation}
u_{t+1}\sim\pi_{\bar\theta}(\cdot\mid h_t,a_t,o_{t+1}),
\end{equation}

\noindent where $\bar\theta$ denotes the policy parameters at trajectory sampling time. RL optimizes the policy based on trajectory rewards, with loss denoted $\mathcal{L}_{\mathrm{RL}}$.

The same trajectories also provide supervision for auxiliary world modeling (WM): the model predicts subsequent environmental observations given history and actions, with observation prediction distribution denoted $p_\theta$. Over the batch observation sample index set $\mathcal{B}_{\mathrm{obs}}$, the full-observation world modeling loss is:

\begin{equation}
\mathcal{L}_{\mathrm{WM}}^{\mathrm{full}}
=
-
\frac{
\sum_{t\in\mathcal{B}_{\mathrm{obs}}}
\log p_\theta(o_{t+1}\mid h_t,a_t)
}{
\sum_{t\in\mathcal{B}_{\mathrm{obs}}}|o_{t+1}|
},
\end{equation}

\noindent where $|o_{t+1}|$ denotes the token count of observation $o_{t+1}$, and the denominator normalizes by the total token count across all observations.

With $\lambda_{\mathrm{WM}}$ controlling auxiliary supervision strength, the joint training objective is:

\begin{equation}
\mathcal{L}_{\mathrm{base}}
=
\mathcal{L}_{\mathrm{RL}}
+
\lambda_{\mathrm{WM}}\mathcal{L}_{\mathrm{WM}}^{\mathrm{full}}.
\end{equation}

\subsection{Decision-Guided Observation Selection}

Complete environmental observations typically mix decision-relevant information with task-irrelevant content. Full-observation world modeling (Eq.~2) treats all content uniformly as world modeling targets, forcing the model to exert equal prediction effort on both noise and useful information. This not only wastes model capacity but may also interfere with the policy's learning of critical information. Our core thesis is: \textbf{only environmental observations that hold value for future decisions merit serving as world modeling targets}. To this end, Decision-Guided Observation Selection uses subsequent policy responses to select supervisory content from observations.

We first partition environmental observation $o_{t+1}$ into a set of candidate units. The partitioning follows a ``minimal semantically complete unit'' principle: for retrieval document feedback, we split at sentence boundaries to obtain single-sentence candidates; for structured webpage observations, we identify semantically complete UI blocks (e.g., product cards, core purchase information, interaction feedback) as candidate units. Implementation details are provided in Appendix~B. Formally, we split the observation into $m_t$ candidate observation units arranged in original order:

\begin{equation}
\mathcal{C}_t=\operatorname{Split}(o_{t+1})
=(c_{t,1},\ldots,c_{t,m_t}).
\end{equation}

For each candidate unit $c_{t,j}$, we retain $h_t$ and $a_t$, replace the current observation with that unit, and use the sampling policy $\pi_{\bar\theta}$ to compute the average token log-likelihood of the true next response under teacher forcing:

\begin{equation}
s_{t,j}
=
\frac{1}{|u_{t+1}|}
\log \pi_{\bar\theta}
\left(u_{t+1}\mid h_t,a_t,c_{t,j}\right).
\end{equation}

We define $s_{t,j}$ as the \textbf{response support score}. Intuitively, $s_{t,j}$ measures the policy's average token log-likelihood for the true subsequent response $u_{t+1}$ given candidate unit $c_{t,j}$. A higher score indicates lower policy decision uncertainty conditioned on that observation unit, meaning the unit contains more information supporting subsequent decision-making. All candidate units are scored against the same true subsequent response.

We perform relative filtering within each observation rather than setting a global absolute threshold. This is because score scales may vary significantly across interactions: some environmental observations have high overall information content, yielding high scores for all candidate units; others contain more noise, resulting in universally lower scores. Relative filtering adapts to the scoring scale of different interactions. Let $\mu_t$ and $\sigma_t$ denote the mean and standard deviation of response support scores across all candidate units in the current observation. The retained set is defined as:

\begin{equation}
\mathcal{C}_t^+
=
\left\{c_{t,j}\;:\;s_{t,j}>\mu_t+\kappa\sigma_t\right\}.
\end{equation}

Parameter $\kappa$ controls filtering strictness. The threshold adjusts with the score distribution of the current observation, making selection criteria adaptive to scoring scales across different interactions. If no candidate unit exceeds the threshold, we skip world modeling for that interaction. Otherwise, we concatenate retained units in original order to form world modeling target $z_{t+1}$:

\begin{equation}
z_{t+1}=\operatorname{Concat}\left(\mathcal{C}_t^+\right).
\end{equation}

For the resulting target, the model predicts its content given interaction history and current action, with average negative log-likelihood:

\begin{equation}
\ell_t
=
-\frac{\log p_\theta(z_{t+1}\mid h_t,a_t)}{|z_{t+1}|}.
\end{equation}

Complete observations remain in the original RL interaction context; filtering only determines auxiliary world modeling target content. However, selecting target content alone is insufficient: for targets the model has already mastered, continued supervision at equal strength wastes training resources; for targets not yet mastered, sufficient learning signal must be maintained. Therefore, we further introduce the Confidence-Adaptive Weighting mechanism.

\subsection{Confidence-Adaptive Weighting}

Target selection determines what feedback content to learn. To avoid continued equal-strength auxiliary supervision on already-mastered targets, Confidence-Adaptive Weighting adjusts supervision strength for each target based on the model's current prediction confidence.

Using the target average negative log-likelihood $\ell_t$ defined in the previous section, we compute prediction confidence:

\begin{equation}
q_t=\exp\left(-\ell_t\right).
\end{equation}

$q_t$ equals the geometric mean of target content token conditional probabilities, reflecting the model's mastery of predicting the entire target. We accordingly compute the adaptive weight:

\begin{equation}
w_t=\operatorname{stopgrad}\left[(1-q_t)^\gamma\right].
\end{equation}

\noindent where $\gamma$ controls confidence weighting decay strength. $\operatorname{stopgrad}$ indicates the weight does not participate in gradient backpropagation; gradients flow through the target prediction loss to update the model.

Each concatenated world modeling target corresponds to one $q_t$ and one $w_t$; all content tokens of that target share this weight. High-confidence targets receive smaller supervision weights, while lower-confidence targets retain stronger learning signal, thereby adaptively adjusting auxiliary supervision pressure according to the model's mastery of the target.

\subsection{Joint Optimization}

Decision-Guided Observation Selection (Section~3.2) determines ``what to predict,'' and Confidence-Adaptive Weighting (Section~3.3) decides ``at what strength to learn.'' The two components jointly constitute our \textbf{Decision-Aware World Modeling Loss}. Let $\mathcal{B}_{\mathrm{WM}}$ denote the index set of interactions in the update minibatch that have valid world modeling targets. The loss is:

\begin{equation}
\begin{aligned}
\mathcal{L}_{\mathrm{WM}}^{\mathrm{DGS}}
&=
\frac{
\sum_{t\in\mathcal{B}_{\mathrm{WM}}}
w_t|z_{t+1}|\ell_t
}{
\sum_{t\in\mathcal{B}_{\mathrm{WM}}}|z_{t+1}|}
\\
&=
-\frac{
\sum_{t\in\mathcal{B}_{\mathrm{WM}}}
w_t\log p_\theta(z_{t+1}\mid h_t,a_t)
}{
\sum_{t\in\mathcal{B}_{\mathrm{WM}}}|z_{t+1}|}.
\end{aligned}
\end{equation}

Compared to $\mathcal{L}_{\mathrm{WM}}^{\mathrm{full}}$, this objective replaces complete observations $o_{t+1}$ with selected $z_{t+1}$ and introduces target-level weights $w_t$. The loss is normalized by the total content token count of selected targets; when $\mathcal{B}_{\mathrm{WM}}$ is empty, the auxiliary loss is set to zero. Combining with the base RL loss yields the complete training objective:

\begin{equation}
\mathcal{L}_{\mathrm{DGS\text{-}WM}}
=
\mathcal{L}_{\mathrm{RL}}
+
\lambda_{\mathrm{WM}}\mathcal{L}_{\mathrm{WM}}^{\mathrm{DGS}}.
\end{equation}

Global coefficient $\lambda_{\mathrm{WM}}$ and target-level weights $w_t$ respectively control the overall strength of auxiliary supervision and individual target contributions.

Algorithm~\ref{alg:dgs-wm} summarizes one sampling and policy update iteration.

\begin{algorithm}[H]
\caption{DGS-WM Training Iteration}
\label{alg:dgs-wm}
\small
\noindent\textbf{Require:} Policy parameters $\theta$; selection threshold $\kappa$; confidence decay $\gamma$; auxiliary coefficient $\lambda_{\mathrm{WM}}$\par
\noindent\textbf{Ensure:} Updated policy parameters $\theta$\par
\vspace{0.2em}
\setcounter{algline}{0}
\algline{$\bar{\theta}\leftarrow\theta$; sample trajectories $\mathcal{D}\sim\pi_{\bar{\theta}}$; initialize $\mathcal{W}\leftarrow\emptyset$}
\algline{\textbf{for each} $(h,a,o,u)\in\mathcal{D}$ \textbf{do}}
\algline[1]{$\mathcal{C}\leftarrow\operatorname{Split}(o)$}
\algline[1]{\textbf{if} $\mathcal{C}\neq\emptyset$ \textbf{then}}
\algline[2]{\textbf{for each} $c\in\mathcal{C}$ \textbf{do}}
\algline[3]{$s(c)\leftarrow\log\pi_{\bar{\theta}}(u\mid h,a,c)/|u|$}
\algline[2]{\textbf{end for}}
\algline[2]{$\mu_t,\sigma_t\leftarrow\operatorname{MeanStd}(\{s(c):c\in\mathcal{C}\})$}
\algline[2]{$\mathcal{C}^{+}\leftarrow\{c\in\mathcal{C}:s(c)>\mu_t+\kappa\sigma_t\}$}
\algline[2]{\textbf{if} $\mathcal{C}^{+}\neq\emptyset$ \textbf{then}}
\algline[3]{$\mathcal{W}\leftarrow\mathcal{W}\cup\{(h,a,\operatorname{Concat}(\mathcal{C}^{+}))\}$}
\algline[2]{\textbf{end if}}
\algline[1]{\textbf{end if}}
\algline{\textbf{end for}}
\algline{\textbf{for each} minibatch $(\mathcal{D}_b,\mathcal{W}_b)$ from $(\mathcal{D},\mathcal{W})$ \textbf{do}}
\algline[1]{\textbf{if} $\mathcal{W}_b=\emptyset$ \textbf{then} $\mathcal{L}_{\mathrm{WM}}\leftarrow 0$}
\algline[1]{\textbf{else}}
\algline[2]{\textbf{for each} $(h,a,z)\in\mathcal{W}_b$ \textbf{do}}
\algline[3]{$\ell_z\leftarrow-\log p_{\theta}(z\mid h,a)/|z|$}
\algline[3]{$q_z\leftarrow\exp(-\ell_z)$; $w_z\leftarrow\operatorname{stopgrad}((1-q_z)^{\gamma})$}
\algline[2]{\textbf{end for}}
\algline[2]{$\mathcal{L}_{\mathrm{WM}}\leftarrow-\frac{\sum_{(h,a,z)\in\mathcal{W}_b}w_z\log p_{\theta}(z\mid h,a)}{\sum_{(h,a,z)\in\mathcal{W}_b}|z|}$}
\algline[1]{\textbf{end if}}
\algline[1]{$\mathcal{J}\leftarrow\mathcal{L}_{\mathrm{RL}}(\mathcal{D}_b)+\lambda_{\mathrm{WM}}\mathcal{L}_{\mathrm{WM}}$}
\algline[1]{$\theta\leftarrow\operatorname{Update}(\theta,\nabla_{\theta}\mathcal{J})$}
\algline{\textbf{end for}}
\end{algorithm}

Candidate scoring and target selection use the sampling policy, while confidence and weights are computed by the current model during updates. At inference time, the model directly generates policy responses based on complete interaction history, requiring no observation filtering or confidence computation, maintaining the same interaction mode as standard agents.

\section{Experiments}

\subsection{Experimental Setup}

\paragraph{Datasets.}
We evaluate on three task categories: (i) \textbf{Multi-hop QA}: HotpotQA \citep{hotpotqa}, 2WikiMultiHopQA \citep{2wikimultihopqa}, MuSiQue \citep{musique}, Bamboogle \citep{bamboogle}; (ii) \textbf{Single-hop QA}: Natural Questions (NQ) \citep{nq}, TriviaQA \citep{triviaqa}, PopQA \citep{popqa}; (iii) \textbf{Interactive Web Shopping}: WebShop \citep{webshop}.

\paragraph{Baselines.}
We compare three method categories: (i) \textbf{Prompting methods}: Direct Inference and ReAct \citep{react}; (ii) \textbf{RL-only methods}: Primary experiments use GRPO \citep{grpo}; cross-algorithm experiments additionally use PPO \citep{ppo}, RLOO \citep{rloo}, and GMPO \citep{gmpo}; (iii) \textbf{RL-augmented world modeling methods}: ECHO \citep{echo} and our proposed DGS-WM, which augment the RL algorithms above with world modeling. ECHO uses complete environmental observations as world modeling targets, serving as a full-observation baseline against DGS-WM's decision-value-guided observation selection. Cross-model and cross-algorithm results compare each RL algorithm alone with its ECHO- and DGS-WM-augmented variants.

\paragraph{Metrics.}
Single-hop and multi-hop QA use Exact Match (EM, \%). Unless otherwise specified, single-hop results are macro-averaged over NQ, TriviaQA, and PopQA; multi-hop results are macro-averaged over HotpotQA, 2WikiMultiHopQA, MuSiQue, and Bamboogle. In Table~\ref{tab:main}, Trivia, 2Wiki, MusiQ, and Bamb abbreviate TriviaQA, 2WikiMultiHopQA, MuSiQue, and Bamboogle, respectively. WebShop reports Success rate (\%) and Score (0--100 scale).

\paragraph{Implementation Details.}
Primary table experiments use Qwen3-1.7B and Qwen3-8B \citep{qwen3}; component ablation uses Qwen3-1.7B. Policy models are fine-tuned via LoRA \citep{lora} with rank 32. For QA tasks, we use the Wiki18 corpus with E5-base-v2 \citep{e5} embeddings and FAISS \citep{faiss} for retrieval. The selection standard deviation coefficient $\kappa$ and confidence decay exponent $\gamma$ are set to 0.5 and 1.0 respectively; following ECHO \citep{echo}, we set the world modeling auxiliary loss coefficient $\lambda_{\mathrm{WM}}=0.02$. All reported results use greedy decoding for evaluation. Additional details are in Appendix~A.

\subsection{Main Results}

\begin{table*}[!t]
\centering
\small
\caption{Main results. QA models train on NQ and HotpotQA; $\dagger$ and $\star$ denote in-domain and out-of-domain QA evaluation. QA reports EM (\%); WebShop reports Success (\%) and Score. Indented rows augment GRPO.}
\label{tab:main}
\setlength{\tabcolsep}{2.5pt}
\begin{tabular*}{\textwidth}{@{\extracolsep{\fill}}l@{\hspace{3pt}}l*{11}{c}@{}}
\toprule
& & \multicolumn{4}{c}{\textbf{Single-hop QA}} & \multicolumn{5}{c}{\textbf{Multi-hop QA}} & \multicolumn{2}{c}{\textbf{WebShop}} \\
\cmidrule(lr){3-6} \cmidrule(lr){7-11} \cmidrule(lr){12-13}
\textbf{Model} & \textbf{Method} & \textbf{NQ}$^{\dagger}$ & \textbf{Trivia}$^{\star}$ & \textbf{PopQA}$^{\star}$ & \textbf{Avg} & \textbf{HotpotQA}$^{\dagger}$ & \textbf{2Wiki}$^{\star}$ & \textbf{MusiQ}$^{\star}$ & \textbf{Bamb}$^{\star}$ & \textbf{Avg} & \textbf{Succ.} & \textbf{Score} \\
\midrule
\multirow{5}{*}{\shortstack{Qwen3-\\1.7B}}
& Direct & 16.18 & 33.99 & 22.12 & 24.10 & 11.03 & 3.75 & 2.03 & 3.20 & 5.00 & 16.60 & 45.51 \\
& ReAct & 18.20 & 36.52 & 22.43 & 25.72 & 12.98 & 11.07 & 4.01 & 20.00 & 12.02 & 17.20 & 54.70 \\
& GRPO & 36.29 & 55.41 & 42.22 & 44.64 & 26.46 & 24.20 & 5.67 & 19.20 & 18.88 & 37.80 & 64.21 \\
& \quad w/ ECHO & 37.09 & 56.71 & 39.16 & 44.32 & 27.08 & 21.06 & 7.49 & 15.20 & 17.71 & 40.40 & 67.93 \\
& \quad \textbf{w/ DGS-WM} & \textbf{37.53} & \textbf{57.46} & \textbf{42.31} & \textbf{45.77} & \textbf{30.30} & \textbf{25.45} & \textbf{7.70} & \textbf{24.80} & \textbf{22.06} & \textbf{43.20} & \textbf{70.39} \\
\midrule
\multirow{5}{*}{\shortstack{Qwen3-\\8B}}
& Direct & 35.57 & 58.03 & 36.74 & 43.45 & 31.72 & 26.66 & 12.54 & 32.80 & 25.93 & 20.20 & 52.12 \\
& ReAct & 32.74 & 59.79 & 34.49 & 42.34 & 35.07 & 39.62 & 13.12 & 39.20 & 31.75 & 23.60 & 59.67 \\
& GRPO & 37.17 & 61.23 & 39.39 & 45.93 & 38.87 & 42.53 & 14.56 & 47.20 & 35.79 & 43.80 & 69.87 \\
& \quad w/ ECHO & 40.25 & 63.47 & 42.52 & 48.75 & 43.63 & 42.64 & 20.73 & 45.60 & 38.15 & 45.00 & 69.32 \\
& \quad \textbf{w/ DGS-WM} & \textbf{43.35} & \textbf{64.39} & \textbf{42.68} & \textbf{50.14} & \textbf{44.85} & \textbf{43.84} & \textbf{22.01} & \textbf{51.20} & \textbf{40.48} & \textbf{47.20} & \textbf{71.75} \\
\bottomrule
\end{tabular*}
\end{table*}

Table~\ref{tab:main} summarizes results across two model scales on single-hop QA, multi-hop QA, and WebShop. DGS-WM consistently outperforms the ECHO co-training baseline across all task groups. On Qwen3-1.7B, DGS-WM achieves 45.77\% and 22.06\% average EM on single-hop and multi-hop QA respectively, improving over ECHO by 1.45\% and 4.35\%. On Qwen3-8B, DGS-WM reaches 50.14\% and 40.48\%, improving over ECHO by 1.39\% and 2.33\%. On WebShop, DGS-WM achieves 43.20\% and 47.20\% success rates on the two models, improving over ECHO by 2.80\% and 2.20\%.

Notably, full-observation world modeling (ECHO) shows unstable performance, sometimes underperforming the pure RL baseline GRPO. For example, on Qwen3-1.7B multi-hop QA, ECHO achieves 17.71\% average EM, \emph{lower} than GRPO's 18.88\% (1.17\% lower); single-hop average EM also drops from 44.64\% to 44.32\%. This suggests that uniformly supervising all environmental observation content as world modeling targets with equal strength may introduce decision-irrelevant noise, interfering with policy optimization. In contrast, DGS-WM consistently improves over GRPO across all settings: on Qwen3-1.7B, single-hop and multi-hop average EM improve by 1.13\% and 3.18\%; on Qwen3-8B, by 4.21\% and 4.69\%. This comparison supports our core thesis: through decision-guided observation selection and confidence-adaptive weighting, DGS-WM provides more stable and effective auxiliary world modeling supervision.

\subsection{Generalization Across Models and Algorithms}

Table~\ref{tab:generalization} validates DGS-WM's generalization advantages over the ECHO-augmented variants. We test three additional models (Qwen2.5-1.5B, Llama-3.2-3B, Qwen2.5-7B) under GRPO, and three additional RL algorithms (PPO, RLOO, GMPO) on Qwen2.5-1.5B. Across all 6 configurations, DGS-WM consistently outperforms both the corresponding RL-only baseline and ECHO-augmented variant. This demonstrates that decision-guided selective world modeling provides stable auxiliary supervision across different model scales and RL algorithms, outperforming full-observation world modeling.

In contrast, the ECHO-augmented variants show inconsistent performance: for GRPO with Llama-3.2-3B or Qwen2.5-7B, they improve single-hop average EM by 1.95\% and 0.79\% over the corresponding RL-only runs; for GRPO with Qwen2.5-1.5B, the single-hop average EM \emph{drops} by 0.70\%, and multi-hop by 0.52\%. With Qwen2.5-1.5B and the three other RL algorithms, the ECHO-augmented variants are also unstable: under PPO, single-hop and multi-hop drop by 0.47\% and 0.94\%; under GMPO, single-hop improves by 2.17\% but multi-hop drops by 0.29\%.

\begin{table}[!t]
\centering
\small
\caption{Generalization across models and RL algorithms. EM (\%) for single-hop and multi-hop QA averaged over all benchmarks in each category. Each row's ``RL-only'' result uses the listed algorithm alone; ``w/ ECHO'' and ``w/ DGS-WM'' are its corresponding augmented variants.}
\label{tab:generalization}
\setlength{\tabcolsep}{2pt}
\begin{tabular}{@{}llrrr@{}}
\toprule
\textbf{Alg.} & \textbf{Model} & \multicolumn{1}{c}{\textbf{RL-only}} & \multicolumn{1}{c}{\textbf{w/ ECHO}} & \multicolumn{1}{c}{\textbf{w/ DGS-WM}} \\
\midrule
\multicolumn{5}{c}{\textit{Single-hop QA}} \\
\midrule
GRPO & Qwen2.5-1.5B & 45.34 & 44.64 & \textbf{46.28} \\
GRPO & Llama-3.2-3B & 47.87 & 49.82 & \textbf{50.75} \\
GRPO & Qwen2.5-7B & 49.76 & 50.55 & \textbf{51.99} \\
PPO & Qwen2.5-1.5B & 44.15 & 43.68 & \textbf{46.06} \\
RLOO & Qwen2.5-1.5B & 44.15 & 44.97 & \textbf{45.85} \\
GMPO & Qwen2.5-1.5B & 42.37 & 44.54 & \textbf{45.54} \\
\midrule
\multicolumn{5}{c}{\textit{Multi-hop QA}} \\
\midrule
GRPO & Qwen2.5-1.5B & 17.22 & 16.70 & \textbf{17.74} \\
GRPO & Llama-3.2-3B & 29.42 & 30.48 & \textbf{33.48} \\
GRPO & Qwen2.5-7B & 32.43 & 34.96 & \textbf{36.54} \\
PPO & Qwen2.5-1.5B & 17.21 & 16.27 & \textbf{17.92} \\
RLOO & Qwen2.5-1.5B & 16.47 & 16.87 & \textbf{17.64} \\
GMPO & Qwen2.5-1.5B & 15.90 & 15.61 & \textbf{17.40} \\
\bottomrule
\end{tabular}
\end{table}

\subsection{Ablation Study}

Table~\ref{tab:ablation} validates the necessity of both DGS-WM components. On Qwen3-1.7B, we compare GRPO alone, its ECHO- and DGS-WM-augmented variants, and two DGS-WM ablations. Removing Decision-Guided Observation Selection (DGOS in Table~\ref{tab:ablation}) reduces DGS-WM to confidence-adaptive weighting on full environmental observations, essentially an enhanced version of ECHO; removing Confidence-Adaptive Weighting (CAW in Table~\ref{tab:ablation}) retains observation selection but fixes all selected target supervision weights to 1 ($\gamma=0$).

Full DGS-WM achieves 45.77\% and 22.06\% average EM on single-hop and multi-hop QA, outperforming all variants. Removing DGOS drops single-hop and multi-hop average EM to 44.45\% and 20.31\%, decreasing by 1.32\% and 1.75\% relative to the full method; this variant uses full observations with confidence weighting, and its performance still exceeds ECHO's direct use of full observations, indicating that confidence-adaptive weighting can partially mitigate noise interference from full observations. Removing CAW drops single-hop and multi-hop average EM to 44.82\% and 21.66\%, decreasing by 0.95\% and 0.40\%, showing that adaptively adjusting supervision strength on top of decision-guided observation selection further improves auxiliary training effectiveness.

\begin{table}[H]
\centering
\small
\caption{Ablation results on Qwen3-1.7B (EM, \% by QA category). The first three rows compare GRPO with its ECHO- and DGS-WM-augmented variants; the final two rows ablate the indicated DGS-WM components.}
\label{tab:ablation}
\setlength{\tabcolsep}{7pt}
\begin{tabular}{@{}lcc@{}}
\toprule
\textbf{Method} & \textbf{Single-hop QA} & \textbf{Multi-hop QA} \\
\midrule
GRPO & 44.64 & 18.88 \\
GRPO w/ ECHO & 44.32 & 17.71 \\
\textbf{GRPO w/ DGS-WM} & \textbf{45.77} & \textbf{22.06} \\
\quad w/o DGOS & 44.45 & 20.31 \\
\quad w/o CAW & 44.82 & 21.66 \\
\bottomrule
\end{tabular}
\end{table}

\subsection{Parameter Sensitivity Analysis}

\begin{figure}[t]
\centering
\includegraphics[width=\columnwidth]{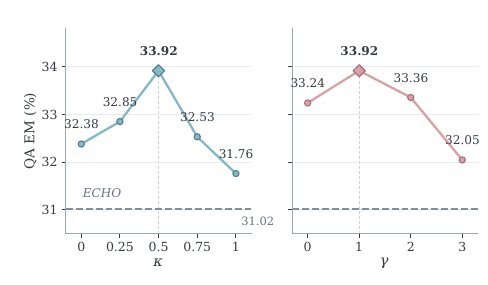}
\caption{Parameter sensitivity on Qwen3-1.7B. Combined QA EM (\%) is the equally weighted average of single-hop and multi-hop averages. Left: varying $\kappa$ with $\gamma=1$ fixed. Right: varying $\gamma$ with $\kappa=0.5$ fixed. The gray dashed line marks the ECHO baseline (31.02\%).}
\label{fig:sensitivity}
\end{figure}

We examine the sensitivity of the selection threshold coefficient $\kappa$ and the confidence decay exponent $\gamma$ on Qwen3-1.7B. As shown in Figure~\ref{fig:sensitivity}, the combined QA average first increases and then decreases with $\kappa$, peaking at $\kappa=0.5$; confidence weighting achieves the best performance at $\gamma=1$.

\section{Conclusion}

We address the problem of non-uniform supervision value in auxiliary world modeling for language agents. Our core thesis is that not all environmental observation content holds equal value for future decisions---only decision-relevant feedback merits serving as world modeling prediction targets. We propose \textbf{DGS-WM} (\textbf{D}ecision-\textbf{G}uided \textbf{S}elective \textbf{W}orld \textbf{M}odeling), which uses \textbf{Decision-Guided Observation Selection} to select supervisory content from observations based on subsequent policy responses, and \textbf{Confidence-Adaptive Weighting} to adaptively adjust supervision strength according to prediction confidence. Across benchmarks spanning single-hop QA, multi-hop QA, and interactive web shopping, DGS-WM achieves consistent improvements over different RL algorithms and model scales, requiring no additional teacher models or human annotations and introducing no inference overhead. Experimental results show that compared to full-observation world modeling methods that may introduce noise interference, selective world modeling provides more stable and effective auxiliary supervision. These results demonstrate that environmental observations in on-policy RL trajectories, after decision-guided selection, provide a simple and practical source of world modeling supervision for language agent training.

\section{Limitations}

Despite its gains across benchmarks, DGS-WM still has two limitations.

First, observation units depend on environment structure: QA uses retrieval-feedback sentences, while WebShop uses page structure. How to identify meaningful units in complex or unstructured feedback remains open; adaptive segmentation based on semantic boundaries, information entropy, or attention is a promising direction.

Second, the response support score estimates an observation's decision value from the policy's average token log-likelihood of the true subsequent response. It assumes that greater consistency with the true response implies greater decision support, but this is correlation, not causation. Counterfactual reasoning or causal interventions could help identify feedback that truly affects decisions.

\clearpage
\bibliography{custom}

\begin{thebibliography}{33}
\providecommand{\natexlab}[1]{#1}

\bibitem[{Ahmadian et~al.(2024)}]{rloo}
Arash Ahmadian et~al. 2024.
\newblock Direct language model alignment from online ai feedback.
\newblock In \emph{arXiv preprint}.

\bibitem[{DeepSeek-AI(2025)}]{deepseek-r1}
DeepSeek-AI. 2025.
\newblock Deepseek-r1: Incentivizing reasoning capability in llms via reinforcement learning.
\newblock \emph{arXiv preprint arXiv:2501.12948}.

\bibitem[{Feng et~al.(2025)Feng, Liu, Cong, and An}]{gigpo}
Lang Feng, Zhenghai Liu, Tingcong Cong, and Bo~An. 2025.
\newblock Group-in-group policy optimization for llm agent training.
\newblock \emph{NeurIPS}.

\bibitem[{Ha and Schmidhuber(2018)}]{world-models}
David Ha and J{\"u}rgen Schmidhuber. 2018.
\newblock World models.
\newblock \emph{arXiv preprint arXiv:1803.10122}.

\bibitem[{Hao et~al.(2023)Hao, Gu, Ma, Hong, Wang, Wang, and Hu}]{reasoning-via-planning}
Shibo Hao, Yi~Gu, Haodi Ma, Joshua~Jiahua Hong, Zhen Wang, Daisy~Zhe Wang, and Zhiting Hu. 2023.
\newblock Reasoning with language model is planning with world model.
\newblock \emph{EMNLP}.

\bibitem[{Ho et~al.(2020)}]{2wikimultihopqa}
Xanh Ho et~al. 2020.
\newblock Constructing a multi-hop qa dataset for comprehensive evaluation of reasoning steps.
\newblock In \emph{COLING}.

\bibitem[{Hu et~al.(2022)Hu, Yelong, Wallis, Allen-Zhu, Li, Wang, Wang, and Chen}]{lora}
Edward~J Hu, Shen Yelong, Phillip Wallis, Zeyuan Allen-Zhu, Yuanzhi Li, Shean Wang, Lu~Wang, and Weizhu Chen. 2022.
\newblock Lora: Low-rank adaptation of large language models.
\newblock In \emph{ICLR}.

\bibitem[{Jin et~al.(2025)Jin, Zeng, Yue, Yoon, Wang, Zumani, and Han}]{search-r1}
Hansi Jin, Zhenrui Zeng, Yinsung Yue, Dong Yoon, Arik, Hamed Wang, Jiawei Zumani, and Jiawei Han. 2025.
\newblock Search-r1: Training search agents with reinforcement learning.
\newblock \emph{ICML}.

\bibitem[{Johnson et~al.(2017)Johnson, Douze, and J{\'e}gou}]{faiss}
Jeff Johnson, Matthijs Douze, and Herv{\'e} J{\'e}gou. 2017.
\newblock Billion-scale similarity search with gpus.
\newblock \emph{IEEE Transactions on Big Data}.

\bibitem[{Joshi et~al.(2017)Joshi, Choi, Weld, and Zettlemoyer}]{triviaqa}
Mandar Joshi, Eunsol Choi, Daniel~S Weld, and Luke Zettlemoyer. 2017.
\newblock Triviaqa: A large scale distantly supervised challenge dataset for reading comprehension.
\newblock In \emph{ACL}.

\bibitem[{Kwiatkowski et~al.(2019)Kwiatkowski, Palomaki, Redfield, Collins, Parikh, Alberti, Epstein, Polosukhin, Devlin, Lee et~al.}]{nq}
Tom Kwiatkowski, Jennimaria Palomaki, Olivia Redfield, Michael Collins, Ankur Parikh, Chris Alberti, Danielle Epstein, Illia Polosukhin, Jacob Devlin, Kenton Lee, et~al. 2019.
\newblock Natural questions: A benchmark for question answering research.
\newblock In \emph{TACL}.

\bibitem[{Li et~al.(2024)Li, Zhao, Wang, Pang, Chen, Quan, Shu, Li, and Li}]{spo}
Guanting Li, Xiaoxiao Zhao, Zijun Wang, Linyao Pang, Yuechen Chen, Chong Quan, Jiaxin Shu, Hanyu Li, and Xingzhang Li. 2024.
\newblock Self-play with execution feedback: Improving instruction-following capabilities of large language models.
\newblock \emph{arXiv preprint arXiv:2406.13542}.

\bibitem[{Li et~al.(2025)Li, Zhao, Liu, and He}]{tree-grpo}
Pandeng Li, Xiangchao Zhao, Yan Liu, and Tianyi He. 2025.
\newblock Tree-search-based policy optimization under stochastic execution.
\newblock \emph{arXiv preprint arXiv:2502.01523}.

\bibitem[{Lightman et~al.(2023)Lightman, Kosaraju, Burda, Edwards, Baker, Lee, Leike, Schulman, Sutskever, and Cobbe}]{process-supervision}
Hunter Lightman, Vineet Kosaraju, Yura Burda, Harri Edwards, Bowen Baker, Teddy Lee, Jan Leike, John Schulman, Ilya Sutskever, and Karl Cobbe. 2023.
\newblock Let's verify step by step.
\newblock \emph{arXiv preprint}.

\bibitem[{Liu et~al.(2026)Liu, Wang, Liu, Guo, and Li}]{imagine-then-plan}
Youwei Liu, Jian Wang, Hanlin Liu, Beichen Guo, and Wenjie Li. 2026.
\newblock Agent learning from adaptive lookahead world models.
\newblock \emph{arXiv preprint arXiv:2601.08955}.

\bibitem[{Lu et~al.(2026)Lu, Lin, Liu, Wu, Lv, Wei, Zhu, Qian, Wang, Chen, Wang, and Tang}]{paw}
Ning Lu, Baijiong Lin, Shengcai Liu, Jiahao Wu, Haoze Lv, Yanbin Wei, Lingting Zhu, Shengju Qian, Xin Wang, Ying-Cong Chen, Qi~Wang, and Ke~Tang. 2026.
\newblock Policy and world modeling co-training for language agents.
\newblock \emph{arXiv preprint arXiv:2606.02388}.

\bibitem[{Mallen et~al.(2023)}]{popqa}
Alex Mallen et~al. 2023.
\newblock Popqa: A dataset for popularity-based question answering.
\newblock In \emph{EMNLP}.

\bibitem[{Meng et~al.(2024)Meng, Zhao, Liu, Fei, Wang, Li, Chen, Wang, Wang, and Chen}]{dapo}
Runsheng Meng, Zigao Zhao, Shuang Liu, Zhaohan Fei, Yiqi Wang, Handong Li, Xinyuan Chen, Mengnan Wang, Yong Wang, and Jiayu Chen. 2024.
\newblock Direct alignment of language models via quality-aware self-refinement.
\newblock \emph{arXiv preprint arXiv:2405.21040}.

\bibitem[{Ouyang et~al.(2022)Ouyang, Wu, Jiang, Almeida, Wainwright, Mishkin, Zhang, Agarwal, Slama, Ray et~al.}]{rlhf}
Long Ouyang, Jeff Wu, Xu~Jiang, Diogo Almeida, Carroll~L Wainwright, Pamela Mishkin, Chong Zhang, Sandhini Agarwal, Katarina Slama, Alex Ray, et~al. 2022.
\newblock Training language models to follow instructions with human feedback.
\newblock \emph{NeurIPS}.

\bibitem[{Press et~al.(2023)}]{bamboogle}
Ofir Press et~al. 2023.
\newblock Measuring and narrowing the compositionality gap in language models.
\newblock In \emph{EMNLP}.

\bibitem[{Schulman et~al.(2017)Schulman, Wolski, Dhariwal, Radford, and Klimov}]{ppo}
John Schulman, Filip Wolski, Prafulla Dhariwal, Alec Radford, and Oleg Klimov. 2017.
\newblock Proximal policy optimization algorithms.
\newblock In \emph{arXiv preprint}.

\bibitem[{Shao et~al.(2024)}]{grpo}
Zhihong Shao et~al. 2024.
\newblock Deepseekmath: Pushing the limits of mathematical reasoning in open language models.
\newblock In \emph{arXiv preprint}.

\bibitem[{Shrivastava et~al.(2026)Shrivastava, Kauffmann, Awadallah, and Papailiopoulos}]{echo}
Vaishnavi Shrivastava, Piero Kauffmann, Ahmed Awadallah, and Dimitris Papailiopoulos. 2026.
\newblock {ECHO}: Terminal agents learn world models for free.
\newblock \emph{arXiv preprint arXiv:2605.24517}.

\bibitem[{Team(2025)}]{qwen3}
Qwen Team. 2025.
\newblock Qwen technical report.
\newblock \emph{arXiv preprint}.

\bibitem[{Trivedi et~al.(2022)}]{musique}
Harsh Trivedi et~al. 2022.
\newblock Musique: Multihop questions via single-hop question composition.
\newblock In \emph{TACL}.

\bibitem[{Wang et~al.(2022)}]{e5}
Liang Wang et~al. 2022.
\newblock Text embeddings by weakly-supervised contrastive pre-training.
\newblock In \emph{arXiv preprint}.

\bibitem[{Williams(1992)}]{reinforce}
Ronald~J Williams. 1992.
\newblock Simple statistical gradient-following algorithms for connectionist reinforcement learning.
\newblock \emph{Machine Learning}, 8:229--256.

\bibitem[{Yang et~al.(2018)Yang, Qi, Zhang, Bengio, Cohen, Salakhutdinov, and Manning}]{hotpotqa}
Zhilin Yang, Peng Qi, Saizheng Zhang, Yoshua Bengio, William~W Cohen, Ruslan Salakhutdinov, and Christopher~D Manning. 2018.
\newblock Hotpotqa: A dataset for diverse, explainable multi-hop question answering.
\newblock In \emph{EMNLP}.

\bibitem[{Yao et~al.(2022)Yao, Chen, Yang, and Narasimhan}]{webshop}
Shunyu Yao, Howard Chen, John Yang, and Karthik Narasimhan. 2022.
\newblock Webshop: Towards scalable real-world web interaction with grounded language agents.
\newblock In \emph{NeurIPS}.

\bibitem[{Yao et~al.(2023)Yao, Zhao, Yu, Du, Shafran, Narasimhan, and Cao}]{react}
Shunyu Yao, Jeffrey Zhao, Dian Yu, Nan Du, Izhak Shafran, Karthik Narasimhan, and Yuan Cao. 2023.
\newblock React: Synergizing reasoning and acting in language models.
\newblock In \emph{ICLR}.

\bibitem[{Yu et~al.(2026)Yu, Peng, Shen, He, Gao, Kim, Gao, and Yu}]{rwml}
Xiao Yu, Baolin Peng, Ruize Shen, Pengcheng He, Suman Gao, Jianyong Kim, Jianfeng Gao, and Zhou Yu. 2026.
\newblock Reinforcement world modeling for llm-based agents.
\newblock \emph{arXiv preprint arXiv:2602.05842}.

\bibitem[{Zhang et~al.(2025)Zhang, Chen, Liu, Xue, Liao, Liu, Wang, Ning, Chen, Xie, Sun, Yang, Zhou, Zhang, Li, Shah et~al.}]{early-experience}
Kai Zhang, Xiangchao Chen, Bo~Liu, Tianci Xue, Yufei Liao, Zhihan Liu, Xiyao Wang, Yuting Ning, Zhaorun Chen, Jian Xie, Yuxuan Sun, Boyu Yang, Jianwei Zhou, Tianyi Zhang, Xia Li, Ashish Shah, et~al. 2025.
\newblock Agent learning via early experience.
\newblock \emph{arXiv preprint arXiv:2510.08558}.

\bibitem[{Zhao et~al.(2025)Zhao, Zhang, Fang, Chen, Wang, Wu, Yuan, Chen, and Yu}]{gmpo}
Peiyi Zhao, Xin Zhang, Yangyi Fang, Jiayin Chen, Tianwei Wang, Wei Wu, Hongyin Yuan, Keqing Chen, and Bowen Yu. 2025.
\newblock Masked thought: Simply masking partial reasoning steps can improve mathematical reasoning learning of language models.
\newblock \emph{arXiv preprint arXiv:2412.03729}.

\end{thebibliography}

\appendix
\section{Implementation Details}
\label{appendix:implementation}

Our experiments cover two types of tasks: Open-Domain Question Answering and Interactive Web Shopping.

\subsection{Datasets}
\label{appendix:datasets}

\subsubsection{Open-Domain QA}

Open-domain QA requires agents to answer questions by interacting with an external retrieval system without pre-given documents. We train on a mixed NQ + HotpotQA dataset and evaluate on multiple benchmarks.

\paragraph{Training Data.}
We mix the following datasets for training:
\begin{itemize}[leftmargin=*,noitemsep,topsep=2pt]
    \item \textbf{Natural Questions (NQ)} \citep{nq}: 79,168 training samples and 8,757 validation samples from real Google search queries
    \item \textbf{HotpotQA} \citep{hotpotqa}: 90,447 training samples and 7,405 validation samples requiring cross-document reasoning
\end{itemize}

From the mixed dataset (169,615 training samples in total), we perform \textbf{stratified sampling} to obtain 27,648 samples for RL training and 1,024 samples for validation. The sampling strategy uses two-level stratification: (1) at the dataset level, NQ and HotpotQA are sampled with weights 1:4; (2) at the category level, within each dataset, samples are drawn proportionally to question categories (single-hop questions for NQ, type and level for HotpotQA). This ensures that the sampled dataset maintains the same distribution of hop counts and question types as the source dataset.

\paragraph{Test Datasets.}
\begin{itemize}[leftmargin=*,noitemsep,topsep=2pt]
    \item \textbf{Natural Questions (NQ)}: 3,610 questions evaluating single-hop fact retrieval
    \item \textbf{TriviaQA} \citep{triviaqa}: 11,313 questions on trivia knowledge
    \item \textbf{PopQA} \citep{popqa}: 14,267 questions focusing on popular culture entities
    \item \textbf{HotpotQA}: 7,405 questions requiring multi-hop reasoning
    \item \textbf{2WikiMultiHopQA} \citep{2wikimultihopqa}: 12,576 questions requiring comparison of two entities
    \item \textbf{MuSiQue} \citep{musique}: 2,417 questions designed to avoid shortcut reasoning
    \item \textbf{Bamboogle} \citep{bamboogle}: 125 manually constructed difficult questions
\end{itemize}

\paragraph{Task Setup.}
Each question allows the agent at most 5 turns to complete the task. In each turn, the agent can execute \texttt{search[query]} (retrieve relevant documents) or \texttt{finish[answer]} (submit the final answer). The retrieval system is based on the \textbf{Wiki18 corpus} (a 2018 snapshot of approximately 5 million Wikipedia articles), using the \textbf{E5-base-v2} encoder \citep{e5} and FAISS index \citep{faiss} for dense retrieval, returning the top-3 documents per search.

\subsubsection{Interactive Web Shopping (WebShop)}

\textbf{WebShop} \citep{webshop} simulates a realistic e-commerce environment with 1.18M products and 12,087 crowd-sourced shopping instructions. The agent needs to browse product pages and complete purchases according to user requirements (e.g., ``buy a waterproof blue backpack under \$50'').

\paragraph{Training Data.}
We use all 10,587 instructions in the training split; 1,000 instructions are held out for validation and 500 for testing.

\paragraph{Task Setup.}
Each shopping task allows at most \textbf{15 steps} to complete. The agent can execute \texttt{search[query]} (search for products), \texttt{click[element]} (click page elements), or \texttt{buy} (purchase item). Success criteria include: (1) purchasing the correct item; (2) item attributes (color, size, etc.) matching requirements; (3) price within budget.

\subsection{Training Details}
\label{appendix:training}

\subsubsection{Model and LoRA Configuration}

\paragraph{Base Models.}
Our main experiments use \textbf{Qwen3-1.7B} and \textbf{Qwen3-8B} \citep{qwen3} as base language models. Cross-model generalization experiments further test \textbf{Qwen2.5-1.5B}, \textbf{Llama-3.2-3B}, and \textbf{Qwen2.5-7B}.

\paragraph{LoRA Fine-tuning.}
To achieve parameter-efficient fine-tuning, we apply LoRA \citep{lora} to all linear projection matrices in attention layers:
\begin{itemize}[leftmargin=*,noitemsep,topsep=2pt]
    \item LoRA rank: $r = 32$
    \item LoRA alpha: $\alpha = 32$
    \item Target modules: Q, K, V, O projections and all linear layers (\texttt{all-linear})
\end{itemize}

\subsubsection{Reinforcement Learning Hyperparameters}

We use \textbf{GRPO} \citep{grpo} as our primary RL algorithm. Key hyperparameters are as follows:

\paragraph{Batch Size and Sampling.}
\begin{itemize}[leftmargin=*,noitemsep,topsep=2pt]
    \item Train batch size: 144
    \item Rollout group size: 8 (8 independent trajectories per group)
    \item Number of groups per rollout: 18 (144 trajectories total)
    \item PPO mini-batch size: 72
    \item PPO epochs: 1
\end{itemize}

\paragraph{Temperature and Sampling Strategy.}
\begin{itemize}[leftmargin=*,noitemsep,topsep=2pt]
    \item Rollout temperature: $T_{\text{rollout}} = 1.0$
    \item Validation temperature: $T_{\text{val}} = 0.0$ (greedy decoding)
\end{itemize}

\paragraph{Optimizer and Learning Rate.}
\begin{itemize}[leftmargin=*,noitemsep,topsep=2pt]
    \item Optimizer: AdamW ($\beta_1 = 0.9, \beta_2 = 0.999, \epsilon = 10^{-8}$)
    \item Learning rate: $5 \times 10^{-6}$
    \item LR schedule: constant (QA); cosine decay (WebShop, 25\% warmup, minimum LR ratio 0.1)
    \item Weight decay: 0.01
    \item Gradient clipping: maximum norm 1.0
\end{itemize}

\subsubsection{DGS-WM Hyperparameters}

The core hyperparameters of our method are:

\paragraph{Decision-Guided Observation Selection.}
\begin{itemize}[leftmargin=*,noitemsep,topsep=2pt]
    \item Selection standard deviation coefficient: $\kappa = 0.5$
    \item Threshold formula: $\tau_t = \mu_t + \kappa \cdot \sigma_t$, so a candidate unit's response support score must exceed the mean by $\kappa$ standard deviations; $\mu_t$ and $\sigma_t$ are the mean and standard deviation of response support scores within the current observation
\end{itemize}

\paragraph{Confidence-Adaptive Weighting.}
\begin{itemize}[leftmargin=*,noitemsep,topsep=2pt]
    \item Confidence decay exponent: $\gamma = 1.0$ (linear decay)
    \item Weight formula: $w_{t,j} = c_t^{\gamma}$, where $c_t$ is the response confidence at turn $t$
\end{itemize}

\paragraph{World Modeling Loss Coefficient.}
\begin{itemize}[leftmargin=*,noitemsep,topsep=2pt]
    \item $\lambda_{\mathrm{WM}} = 0.02$ (consistent with ECHO)
\end{itemize}

\subsubsection{Reward Function}

We adopt sparse binary reward signals for all tasks:

\paragraph{QA Task Rewards.}
\begin{itemize}[leftmargin=*,noitemsep,topsep=2pt]
    \item Success (exact answer match): $R = 1$
    \item Failure (answer mismatch or exceeding turn limit): $R = 0$
    \item Invalid action penalty: $-0.01$ per invalid action (applied to corresponding action tokens in per-action mode)
\end{itemize}

\paragraph{WebShop Task Rewards.}
\begin{itemize}[leftmargin=*,noitemsep,topsep=2pt]
    \item Success (correct item purchased with all attributes fully matched): $R = 1$
    \item Failure (no purchase, wrong item, or attributes not fully matched): $R = 0$
    \item Invalid action penalty: $-0.01$ per invalid action (applied to corresponding action tokens in per-action mode)
\end{itemize}

\section{Observation Unit Partitioning}
\label{appendix:observation_units}

\subsection{QA Tasks: Sentence Splitting}

For retrieval document feedback, we split at sentence boundaries using forward sentence splitting with the following rules:

\textbf{Split pattern:} We use the regex pattern \texttt{(?<=[.!?])\textbackslash s+(?=[A-Z0-9"'\textbackslash(\textbackslash[])}, which splits after sentence-ending punctuation followed by whitespace and a capital letter or digit.

\textbf{Abbreviation protection:} Common abbreviations (Dr., U.S., Mr., Mrs., Ph.D., etc.) are protected from splits by checking whether the period is preceded by these patterns.

\textbf{Single-letter protection:} Single capital letters followed by periods (e.g., "A. Smith") are preserved as part of the same sentence to avoid breaking names or list markers.

Each resulting sentence becomes one candidate observation unit $c_{t,j}$ with metadata tracking its paragraph ID, rank within the paragraph, sentence index, and character positions in the original observation.

\subsection{WebShop: Semantic UI Block Identification}

For webpage observations, we identify semantic blocks based on page type inference:

\textbf{Page type inference:} We classify each page as one of: \texttt{search\_results}, \texttt{product\_main}, \texttt{search\_home}, or \texttt{done} based on page structure markers (search results list, product details section, search bar, confirmation message).

\textbf{Semantic block construction by page type:}

\begin{itemize}[noitemsep]
\item \textbf{search\_results page:} Each product card becomes one block, containing ASIN, description, and price as a unit.
\item \textbf{product\_main page:} Core purchase information (selected options + title + price + rating + "Buy Now" button) forms one block; product description paragraphs are split into sentences; customer review blocks remain intact.
\item \textbf{search\_home page:} Search bar and category buttons each form separate blocks.
\item \textbf{Feedback blocks:} System messages (e.g., "Added to cart") remain as complete units.
\end{itemize}

\textbf{Fallback handling:} For unrecognized page structures, interactive buttons remain as whole-line units, while plain text content is split into sentences using the QA sentence splitting rules.

\subsection{Granularity Design Rationale}

The "minimal semantically complete unit" principle avoids two failure modes:

\textbf{Over-coarse splitting} (e.g., entire paragraphs or full product pages): Dilutes response support scores by mixing decision-relevant content with irrelevant context, reducing selection precision.

\textbf{Over-fine splitting} (e.g., individual words or tokens): Breaks semantic integrity, making isolated fragments uninformative even when the complete unit would support decisions.

Our task-aware unit definitions balance these concerns: sentences preserve propositional content in documents, while UI blocks maintain functional coherence in interactive environments.


\definecolor{caseink}{HTML}{263747}
\definecolor{caseselection}{HTML}{1B634F}
\definecolor{caseborder}{HTML}{CFD7DE}
\definecolor{casebackground}{HTML}{F6F8FA}
\definecolor{caseresponse}{HTML}{EEF5F1}
\newcommand{\caseselected}[1]{\textcolor{caseselection}{\textbf{#1}}}
\newcommand{\casedocument}[2]{%
  {\sffamily\color{caseink}#1}\quad Title: #2\par\smallskip}
\newtcolorbox{caseobservation}[2][]{%
  colback=white,colframe=caseborder,colbacktitle=casebackground,
  coltitle=caseink,fonttitle=\sffamily\bfseries,
  fontupper=\normalsize,boxrule=0.45pt,arc=1.5pt,
  left=8pt,right=8pt,top=7pt,bottom=7pt,
  toptitle=5pt,bottomtitle=5pt,
  before skip=8pt,after skip=8pt,
  title={#2},#1}
\newtcolorbox{caseoutcome}{%
  colback=caseresponse,colframe=caseborder,boxrule=0.45pt,arc=1.5pt,
  left=8pt,right=8pt,top=6pt,bottom=6pt,
  fontupper=\normalsize,before skip=7pt,after skip=8pt}
\newcommand{\caseresult}[2]{%
  \begin{caseoutcome}
  \textbf{Selection:} #1\par\smallskip
  \textbf{Subsequent policy response:}\quad\texttt{#2}\hfill\ding{51}
  \end{caseoutcome}}

\clearpage
\twocolumn[{%
\begin{minipage}{\textwidth}
\section{Case Studies}
\label{appendix:case_studies}
This section presents examples of DGS-WM's observation selection compiled from real training logs.
\textbf{Bold text}, additionally highlighted in dark green within the observations, marks content selected as WM prediction targets.

\subsection{Single-Hop Question Answering (Qwen3-1.7B)}
\label{appendix:case_single_hop}
\noindent\textbf{Question:} What became the standard German term for the art song?
\begin{caseobservation}{Environment observation: 3 retrieved documents, 18 candidate sentences}
\casedocument{[Document 10565234; rank 1]}{Art song}
\noindent Text: wordless vocalise written by a classical composer is sometimes considered an art song and sometimes not. Other factors help define art songs: Art songs have been composed in many languages, and are known by several names. \caseselected{The German tradition of art song composition is perhaps the most prominent one; it is known as "Lieder".} In France, the term "mélodie" distinguishes art songs from other French vocal pieces referred to as chansons. The Spanish canción and the Italian canzone refer to songs generally and not specifically to art songs. The composer's musical language and interpretation of the text often dictate the\par
\medskip
\casedocument{[Document 12808472; rank 2]}{Song}
\noindent Text: traditions. German-speaking communities use the term art song ("Kunstlied") to distinguish so-called "serious" compositions from folk song ("Volkslied"). The lyrics are often written by a poet or lyricist and the music separately by a composer. \caseselected{Art songs may be more formally complicated than popular or folk songs, though many early Lieder by the likes of Franz Schubert are in simple strophic form.} The accompaniment of European art songs is considered as an important part of the composition. Some art songs are so revered that they take on characteristics of national identification. Art songs emerge from the tradition of singing romantic\par
\medskip
\casedocument{[Document 12168286; rank 3]}{Art song}
\noindent Text: Art song An art song is a vocal music composition, usually written for one voice with piano accompancement, and usually in the classical art music tradition. By extension, the term "art song" is used to refer to the collective genre of such songs (e.g., the "art song repertoire"). An art song is most often a musical setting of an independent poem or text, "intended for the concert repertory" "as part of a recital or other relatively formal social occasion". While many pieces of vocal music are easily recognized as art songs, others are more difficult to categorize. For example, a\par
\end{caseobservation}
\caseresult{18 candidate sentences $\rightarrow$ \textbf{2 selected sentences (11.1\%)}}{<answer>Lieder</answer>}
\end{minipage}%
}]
\clearpage
\twocolumn[{%
\begin{minipage}{\textwidth}
\subsection{Multi-Hop Question Answering (Qwen3-8B)}
\label{appendix:case_multi_hop}
\noindent\textbf{Question:} The director of the romantic comedy "Big Stone Gap" is based in what New York city?
\begin{caseobservation}{Environment observation: 3 retrieved documents, 27 candidate sentences}
\casedocument{[Document 11856191; rank 1]}{Big Stone Gap (film)}
\noindent Text: \caseselected{Big Stone Gap is a 2014 American drama romantic comedy film written and directed by Adriana Trigiani and produced by Donna Gigliotti} for Altar Identity Studios, a subsidiary of Media Society. Based on Trigiani's 2000 best-selling novel of the same name, the story is set in the actual Virginia town of Big Stone Gap circa 1970s. The film had its world premiere at the Virginia Film Festival on November 6, 2014.\par
\medskip
\casedocument{[Document 21947720; rank 2]}{Adriana Trigiani}
\noindent Text: Adriana Trigiani is an Italian American best-selling author of eighteen books, television writer, film director and entrepreneur \caseselected{based in Greenwich Village, New York City.} She is also a published playwright. Trigiani has published a novel a year since 2000. Her tenth novel, "The Shoemaker's Wife," her first historical novel, was published March 27, 2012, by Harper Collins. It debuted at number one on the "New York Times" Bestseller list and remained on the list for five weeks.\par
\medskip
\casedocument{[Document 8647293; rank 3]}{Big Stone Gap, Virginia}
\noindent Text: Big Stone Gap is a town in Wise County, Virginia, United States. The population was 5,643 at the 2010 census. The town is home to a campus of Mountain Empire Community College. It is also home to author and film producer Adriana Trigiani who wrote and filmed the movie "Big Stone Gap" based on her novel of the same name.\par
\medskip
\end{caseobservation}
\caseresult{27 candidate sentences $\rightarrow$ \textbf{2 selected sentences (7.4\%)}}{<answer>Greenwich Village</answer>}
\noindent\textbf{Interpretation.} This question requires reasoning across documents: the first document identifies Adriana Trigiani as the director, and the second gives her location. DGS-WM selects only these two key sentences.
\end{minipage}%
}]
\clearpage
\twocolumn[{%
\begin{minipage}{\textwidth}
\subsection{Product Search (Qwen3-8B, WebShop)}
\label{appendix:case_product_search}
\noindent\textbf{Task:}
\textit{i need a long lasting fragrance gift set for women, and price lower than 80.00 dollars}
\par\smallskip
\noindent\textbf{Current step:} Select a suitable product from the search results to click.
\begin{caseobservation}{Search results}
\textit{10 products, each with a title, price, and brief description; approximately 50 candidate UI blocks.}
\par\medskip
[Product 1]\\
Floral Summer Eau de Toilette 100ml\\
\$45.99\\
Light floral scent, perfect for daily wear\par
\smallskip
\caseselected{[Product 2]}\\
\caseselected{Luxury Fragrance Gift Set for Women}\\
\caseselected{\$64.99}\\
\caseselected{Long-lasting Eau de Parfum + Body Lotion + Travel Spray}\par
\smallskip
[Product 3]\\
Classic Rose Perfume 50ml\\
\$89.00\\
Premium rose essence, elegant bottle design\par
\smallskip
[Product 4]\\
Travel Mini Perfume Collection\\
\$29.99\\
5 x 10ml mini bottles, various scents\par
\smallskip
\caseselected{[Product 5]}\\
\caseselected{Premium Women's Fragrance Duo}\\
\caseselected{\$72.50}\\
\caseselected{Lasting scent, gift-ready packaging}\par
\smallskip
[Product 6]\\
Fresh Ocean Breeze Spray 200ml\\
\$38.00\\
Refreshing aquatic scent\par
\smallskip
[4 other products omitted...]\par
\end{caseobservation}
\caseresult{Approximately 50 UI blocks $\rightarrow$ \textbf{6 selected (12\%)}}{<click>Product 2</click>}
\noindent\textbf{Interpretation.} The agent selects a product that satisfies all task constraints: a long-lasting fragrance gift set for women priced below \$80.
\end{minipage}%
}]
\clearpage
\twocolumn[{%
\begin{minipage}{\textwidth}
\subsection{Product-Detail Confirmation (Qwen3-8B, WebShop)}
\label{appendix:case_product_details}
\noindent\textbf{Task:}
\textit{i need a long lasting fragrance gift set for women, and price lower than 80.00 dollars}
\par\smallskip
\noindent\textbf{Current step:} After clicking the product, check its details against the task requirements.
\begin{caseobservation}{Product details: 64 candidate UI blocks}
\begin{minipage}[t]{0.485\linewidth}
\raggedright
\caseselected{Product Details:}\par
\smallskip
\caseselected{[Core Info]}\\
\caseselected{Name: Luxury Fragrance Gift Set for Women}\\
\caseselected{Price: \$64.99}\\
\caseselected{Set Includes: Eau de Parfum 50ml + Body Lotion 75ml + Travel Spray 10ml}\par
\smallskip
[Shipping]\\
Free shipping on orders over \$50\\
Estimated delivery: 3-5 business days\\
30-day return policy\par
\smallskip
\caseselected{[Customer Reviews - Top Rated]}\\
\caseselected{\ding{72}\ding{72}\ding{72}\ding{72}\ding{72} "Lasts all day! Great for gifting."}\\
\caseselected{\ding{72}\ding{72}\ding{72}\ding{72}\ding{72} "Perfect gift set, my mom loved it."}\\
\ding{72}\ding{72}\ding{72}\ding{73}\ding{73} "Nice scent but bottle design could be better."\par
\smallskip
[Brand Story]\\
Founded in 1985, our brand has been crafting luxury fragrances for over 35 years. Each fragrance is carefully formulated using premium ingredients sourced from around the world. Our master perfumers blend traditional techniques with modern innovation...\par
\end{minipage}\hfill%
\begin{minipage}[t]{0.465\linewidth}
\raggedright
[Product Specifications]\\
Fragrance Family: Floral Oriental\\
Top Notes: Bergamot, Mandarin\\
Heart Notes: Jasmine, Rose\\
Base Notes: Vanilla, Musk\\
Bottle Material: Glass with gold-tone cap\\
Made in France\par
\smallskip
\caseselected{[Key Features]}\\
\caseselected{\textbullet{} Long-lasting formula (8-10 hours)}\\
\textbullet{} Gift-ready elegant packaging\\
\textbullet{} Suitable for all occasions\par
\smallskip
[Recommendations]\\
You may also like:\\
- Similar Perfume Set (\$79.99)\\
- Luxury Body Care Collection (\$54.99)\\
- Travel Size Sampler (\$29.99)\par
\smallskip
[Size Options]\\
\textbullet{} 50ml Set - \$64.99 (Selected)\\
\textbullet{} 100ml Set - \$94.99\\
\textbullet{} Travel Size Only - \$24.99\par
\end{minipage}
\end{caseobservation}
\caseresult{64 UI blocks $\rightarrow$ \textbf{8 selected (12.5\%)}}{<buy>}
\noindent\textbf{Interpretation.} The agent focuses on information needed to verify the task requirements (set contents, longevity, and price), ignoring secondary content such as the brand story, ingredient details, and recommended products.
\par\medskip
\noindent\textbf{Summary.} These cases illustrate how DGS-WM selectively focuses on decision-relevant observations across tasks, substantially reducing the scope of world-modeling supervision (selection rates of 7.4--12.5\%) while preserving correct policy responses.
\end{minipage}%
}]

\end{document}